\documentclass[11pt]{cambrian}

\usepackage{booktabs}       % professional-quality tables
\usepackage{amsfonts}       % blackboard math symbols
\usepackage{nicefrac}       % compact symbols for 1/2, etc.
\usepackage{microtype}      % microtypography
\usepackage{multirow}       % for multirow cells in tables
\definecolor{bluelink}{RGB}{0,113,188}
\definecolor{greenlink}{RGB}{0,188,113}
\hypersetup{
    colorlinks=true,%
    citecolor=green!93!black,%
    filecolor=redlink,%
    linkcolor=red!93!black,%
    urlcolor=bluelink
}

\usepackage[ruled,vlined,noend]{algorithm2e}
\usepackage{subcaption}
\usepackage{xspace}

\usepackage[normalem]{ulem} % for \uline

\usepackage[retain-explicit-plus]{siunitx}

\usepackage{titlesec}
\titlespacing*{\paragraph}
    {0pt}     % left margin
    {0.25em}  % space before (vertical above)
    {1em}  % space after (vertical below, adjust as needed)

\usepackage{wrapfig}
\usepackage[export]{adjustbox}% for valign

\definecolor{rowheader}{RGB}{240,240,240}
\definecolor{oursrowheader}{RGB}{200,220,255}
\definecolor{ours}{RGB}{220,240,255}
\definecolor{inccol}{RGB}{245,245,245}
\definecolor{inctext}{RGB}{100,100,100}
\definecolor{pos}{RGB}{0,128,0}
\definecolor{neg}{RGB}{200,0,0}
\definecolor{navyblue}{HTML}{0071BC}

\newcommand{\p}[1]{{\color{pos}{#1}}}
\newcommand{\n}[1]{{\color{neg}{#1}}}

\newcommand{\displaytodo}[1]{} % uncomment to hide todos

\newcommand{\dataset}{\texttt{SIMS-VSI}\xspace}
\newcommand{\framework}{\texttt{SIMS-V}\xspace}

\title{
    \vspace{-1em}
    \centering
    \Large \framework{}: Simulated Instruction-Tuning for Spatial Video Understanding
}

\author{
    \vspace{-0.25em}
    Ellis Brown\textsuperscript{1} \quad
    Arijit Ray\textsuperscript{2} \quad
    Ranjay Krishna\textsuperscript{3} \quad
    Ross Girshick\textsuperscript{4} \quad
    Rob Fergus\textsuperscript{1} \quad
    Saining Xie\textsuperscript{1} \\
    \vspace{-0.25em}
    \textsuperscript{1}New York University \quad
    \textsuperscript{2}Boston University \quad
    \textsuperscript{3}AllenAI \quad
    \textsuperscript{4}Vercept

    {
        \vspace{-0.25em}
        \normalsize
        \url{https://ellisbrown.github.io/sims-v}
    }
    \vspace{-0.5em}
}

\begin{abstract}
    \vspace{-0.5em}
    Despite impressive high-level video comprehension, multimodal language models struggle with spatial reasoning across time and space.
    While current spatial training approaches rely on real-world video data, obtaining diverse footage with precise spatial annotations remains a bottleneck.
    To alleviate this bottleneck, we present \framework{}---a systematic data-generation framework that leverages the privileged information of 3D simulators to create spatially-rich video training data for multimodal language models.
    Using this framework, we investigate which properties of simulated data drive effective real-world transfer through systematic ablations of question types, mixes, and scales.
    We identify a minimal set of three question categories (metric measurement, perspective-dependent reasoning, and temporal tracking) that prove most effective for developing transferable spatial intelligence, outperforming comprehensive coverage despite using fewer question types.
    These insights enable highly efficient training: our 7B-parameter video LLM fine-tuned on just 25K simulated examples outperforms the larger 72B baseline and achieves competitive performance with proprietary models on rigorous real-world spatial reasoning benchmarks.
    Our approach demonstrates robust generalization, maintaining performance on general video understanding while showing substantial improvements on embodied and real-world spatial tasks.
\end{abstract}

\begin{document}

\maketitle

\begin{figure*}[h!]
    \vspace{1em}
    \centering
    \includegraphics[width=\linewidth]{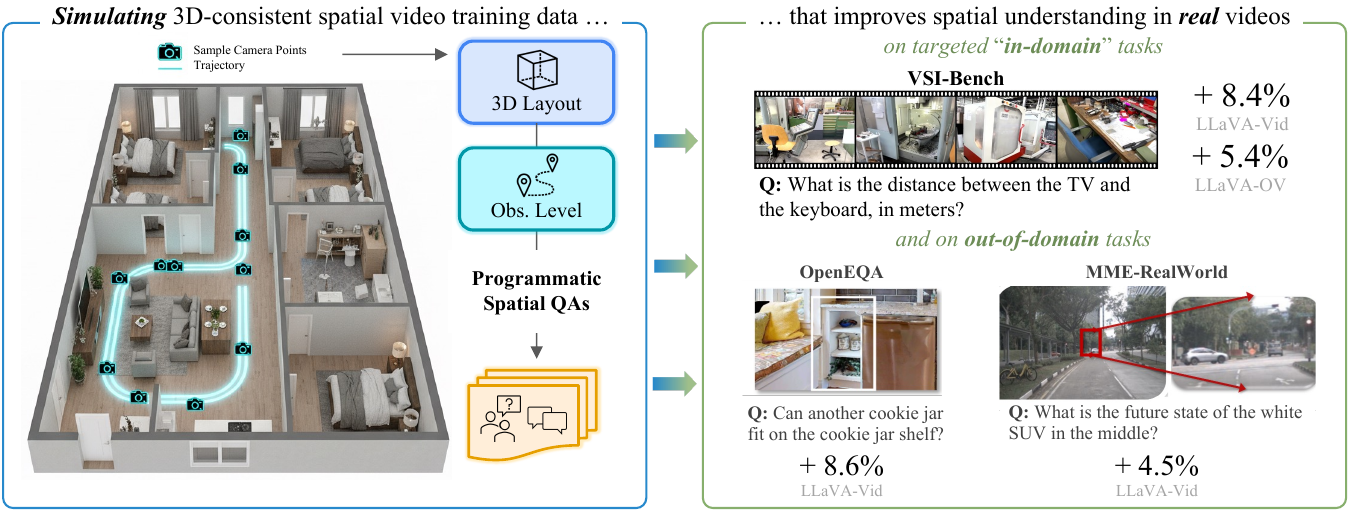}
    \caption{
        \textbf{\framework{} enables learning real-world spatial concepts in simulation.}
        We generate spatially-rich videos with dense spatial annotations via privileged simulator data, creating diverse question-answer pairs.
        Models trained on this simulated data transfer effectively to real-world spatial reasoning benchmarks.
    }
\label{fig:teaser}
\end{figure*}

\clearpage

\section{Introduction}

Despite strong performance on most benchmarks~\cite{zhang2024llavanextvideo, liu2023llava, openai2023gpt, team2023gemini, zhu2024minigpt}, multimodal large language models (MLLMs) struggle with understanding spatial reasoning~\cite{ma2023crepe,chen2024spatialvlm, ray2025sat, fu2024blink, tong2025cambrian, hsieh2023sugarcrepe}.
This challenge is particularly acute for spatiotemporal reasoning in videos, where models must track and reason about spatial configurations across time~\cite{yang2024think}.
Many attribute this weakness to a lack of spatial annotations in web-scale image-text pretraining data~\cite{kakaobrain2022coyo-700m, schuhmann2022laionb, yang2024think, ray2025sat, cheng2025spatialrgpt}.
While real-world video data with precise 3D spatial annotations would address this gap, such data is prohibitively expensive to collect at scale.
Simulators offer a compelling alternative with perfect ground truth and scalable generation, but sim-to-real transfer remains a fundamental challenge in computer vision~\cite{tobin2017domain, matas2018sim, tzeng2020adapting, liu2020stereogan}.
Critical questions remain: \emph{What properties of simulated spatial data are necessary for effective transfer to real-world video understanding? Which question types and data configurations matter most?}

To investigate these questions, we present \framework{}, a \uline{s}imulated \uline{i}nstruction-tuning framework for \uline{m}ultimodal \uline{s}patial \uline{v}ideo understanding.
\framework{} leverages procedurally generated 3D environments and the privileged information available in simulators to programmatically create rich question-answer pairs about spatial configurations across videos.
As a concrete instantiation, we generate \dataset{}, a dataset comprising over 200k spatial question-answer pairs across 2.5k video trajectories spanning 1.2k diverse indoor scenes.
Unlike prior work using simulation for single-image spatial tasks~\cite{ray2025sat}, we formulate complex questions requiring spatiotemporal reasoning---perspective-taking, metric distance estimation, and temporal tracking of object appearances across minutes-long video trajectories.
This controlled framework enables us to systematically investigate which properties of simulated data drive effective sim-to-real transfer.

Through controlled experiments, we uncover key insights about designing effective simulated spatial training data.
We identify a minimal set of three question categories (metric measurement, perspective-taking, and spatiotemporal tracking) that prove most effective for developing transferable spatial intelligence.
Notably, this focused mix outperforms a comprehensive training set mirroring the full VSI-Bench distribution, demonstrating that carefully selected supervision on core spatial reasoning dimensions can be sufficient---and even superior---to comprehensive coverage.
Additionally, we find that spatial reasoning capabilities can be learned remarkably efficiently: training on just 5K targeted simulated examples enables performance competitive with models trained on orders of magnitude more data.

Applying these insights, we demonstrate that a 7B-parameter LLaVA model fine-tuned on just 25K simulated examples achieves competitive performance with large proprietary models like GPT-4o~\cite{hurst2024gpt} and Gemini-1.5-Pro~\cite{team2024gemini}.
To ensure rigorous evaluation of genuine visual reasoning capabilities, we report results on both VSI-Bench~\cite{yang2024think} and its robustified counterpart VSI-Bench-Debiased~\cite{brown2025benchmark} which is refined to minimize non-visual shortcuts.
Our approach also demonstrates strong generalization, maintaining performance on general video understanding benchmarks while showing substantial improvements on diverse spatial tasks including embodied reasoning~\cite{OpenEQA2023} and real-world scenarios~\cite{fu2025videomme, mangalam2023egoschema}.

The remainder of this paper is organized as follows:
We first situate our work within related research on spatial reasoning, video-language models, and synthetic data (\cref{sec:related_work}).
We then detail the \framework{} framework for generating controlled simulated spatial training data (\cref{sec:method}).
Next, we present systematic experiments investigating which question types, data mixes, and quantities drive effective real-world transfer (\cref{sec:investigating}).
We evaluate transfer of understanding to real-world benchmarks (\cref{sec:evaluation}).
Finally, we discuss limitations and future directions (\cref{sec:discussion}).

\section{Simulated Spatial Instruction-Tuning}\label{sec:method}
\begin{figure*}[t]
    \centering
    \includegraphics[width=\linewidth]{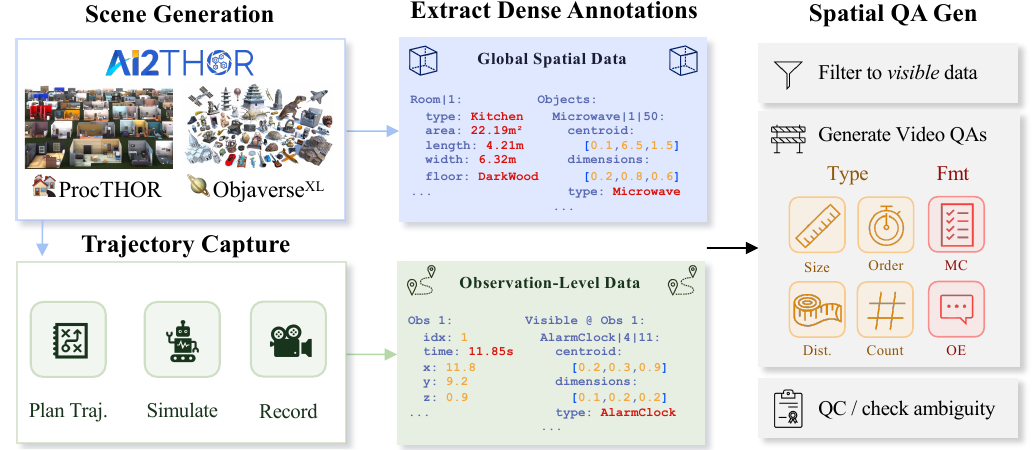}
    \caption{
        \textbf{The \framework{} pipeline generates diverse spatial training data with perfect ground truth.}
        We procedurally generate 3D scenes using AI2-THOR~\cite{kolve2017ai2}, ProcTHOR~\cite{deitke2022procthor}, and Objaverse~\cite{deitke2023objaverse}, capture agent navigation trajectories, extract dense annotations (global spatial layout and per-frame observations), and programmatically generate quality-controlled question-answer pairs spanning diverse spatial reasoning categories.
        This systematic pipeline enables controlled ablations of question types and data configurations, maintaining perfect spatial ground truth.
    }\label{fig:method-overview}
\end{figure*}

Simulators offer a uniquely powerful tool for addressing this data bottleneck.
Unlike real-world footage, simulators provide \emph{perfect} ground truth spatial and per-observation information while enabling systematic control over scene complexity and camera trajectories.
This combination allows us to investigate the minimal requirements for learning transferable spatial reasoning capabilities.
Using this framework, we generate \dataset{}, a dataset comprising over 200k question-answer pairs across 2.5k video trajectories through 1.2k diverse indoor scenes.
In this section, we describe our Simulated Instruction-tuning for Multimodal Spatial-understanding (\framework{}) framework for generating high-quality spatial training data.

\subsection{Simulated Video Generation}

To create simulated instruction-tuning data, we first generate realistic indoor environments and capture trajectories of agents navigating through these spaces, carefully extracting comprehensive metadata at each step.

\paragraph{Scene Generation and Trajectory Capture}
We leverage the AI2-THOR simulator~\cite{kolve2017ai2} with environments procedurally generated by ProcTHOR~\cite{deitke2022procthor} using objects from Objaverse~\cite{deitke2023objaverse} for added diversity.
Each generated environment contains approximately 30--50 objects across 3--8 rooms, providing complex spatial arrangements necessary for challenging reasoning tasks.

We capture multiple trajectories through each environment, touring all rooms from various starting points to ensure comprehensive spatial coverage.
To accomplish this, we plan a shortest path traversal to the center of each room, followed by a 360-degree rotation at the center to capture a full view of the room.
Videos are captured at 10 frames per second with a resolution of 680 \texttimes{} 384, resulting in sequences of 60--1,800 frames (12s--3min) per trajectory.
We render scenes in the highest quality available via the simulator by default.

\paragraph{Dense Annotation Extraction}

Our pipeline extracts two complementary metadata types:

\noindent\emph{Observation-level data}.\quad Per-frame information including visible objects, instance segmentation masks, and agent position \& viewpoint. This captures all information about what artifacts are visible at each time step.

\noindent\emph{Global spatial data}.\quad
Comprehensive scene-level information including room layouts, 3D object positions, and object metadata. This provides perfect ground truth for generating spatial questions even about objects currently outside the agent's field of view.

\subsection{Spatial Question-Answer Generation}

Using our dense metadata, we programmatically generate precise spatial reasoning questions with guaranteed ground truth answers. This automated process leverages both observation-level and global spatial metadata to create questions that would be prohibitively expensive to annotate in real-world video data. Specifically, our question generation pipeline follows these steps:

\begin{enumerate}[itemsep=0.125em,topsep=0pt,leftmargin=16pt]
    \item \emph{Metadata preprocessing}: We first identify ``salient'' objects (those visible above a minimum pixel threshold) and extract their spatial relationships, temporal appearances, and 3D positions.
    \item \emph{Question formulation}: Based on the preprocessed metadata, we generate questions across diverse spatial reasoning categories, such as layout understanding (counting, measurements, distances), relative spatial relationships (positional and directional), and temporal-spatial memory (appearance order).
    \item \emph{Format variation}: For each question, we automatically generate both open-ended and multiple-choice versions, with carefully balanced multiple-choice distractors.
    \item \emph{Quality control}: The generation process incorporates rigorous quality control, including minimum visibility thresholds for objects, distance requirements between objects, and angle thresholds for directional questions. This ensures that all generated questions are clearly answerable from the video content without ambiguity.
\end{enumerate}

\noindent
This systematic data generation framework enables us to conduct controlled experiments investigating which properties of simulated data drive effective transfer to real-world spatial reasoning, which we present in the following sections.

\begin{figure*}[t]
    \centering
    \includegraphics[width=\linewidth]{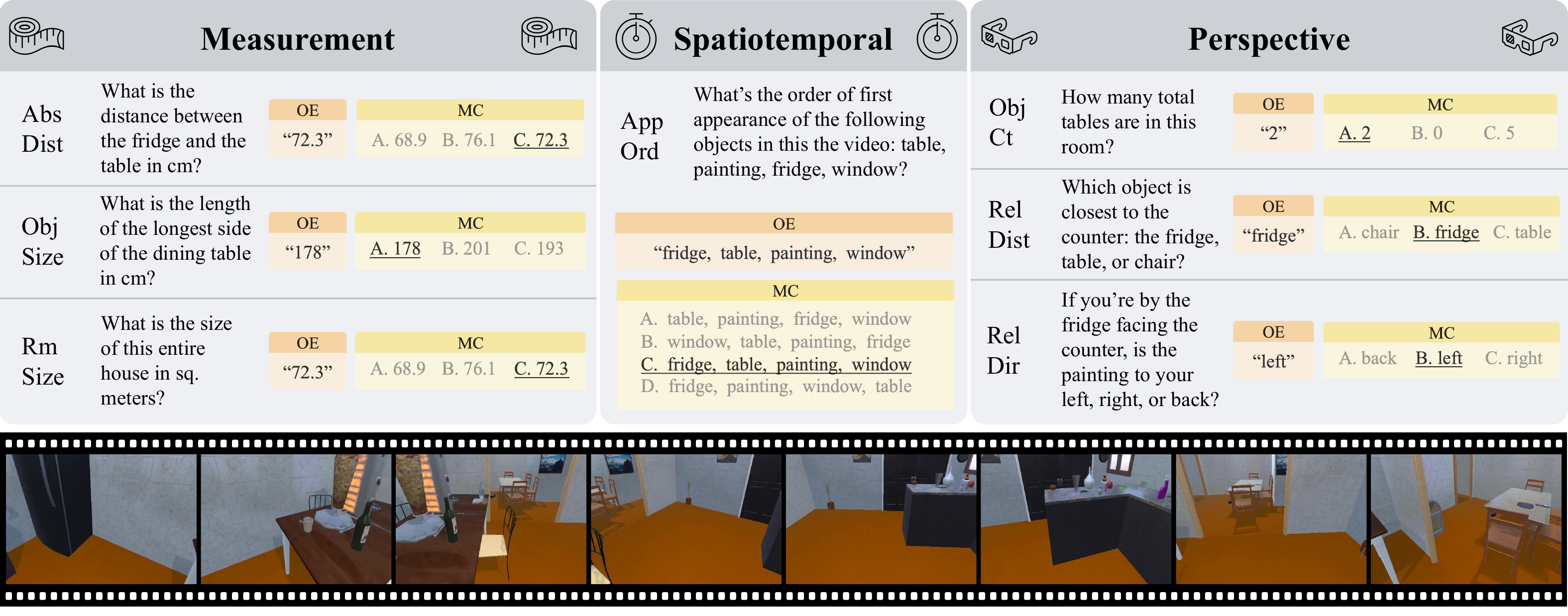}
    \caption{
        Examples of different question types used in our experiments. Each question is shown alongside its corresponding visual context from a simulated environment. The questions span diverse spatial reasoning capabilities including numerical measurement, relative positioning, and temporal tracking. Full details of all question formats are provided in \cref{app:question_templates}.
    }\label{fig:question_type_visualization}
\end{figure*}

\section{Investigating Simulated Data Properties}\label{sec:investigating}

\begin{figure*}[t]
    \centering
    \includegraphics[width=\linewidth]{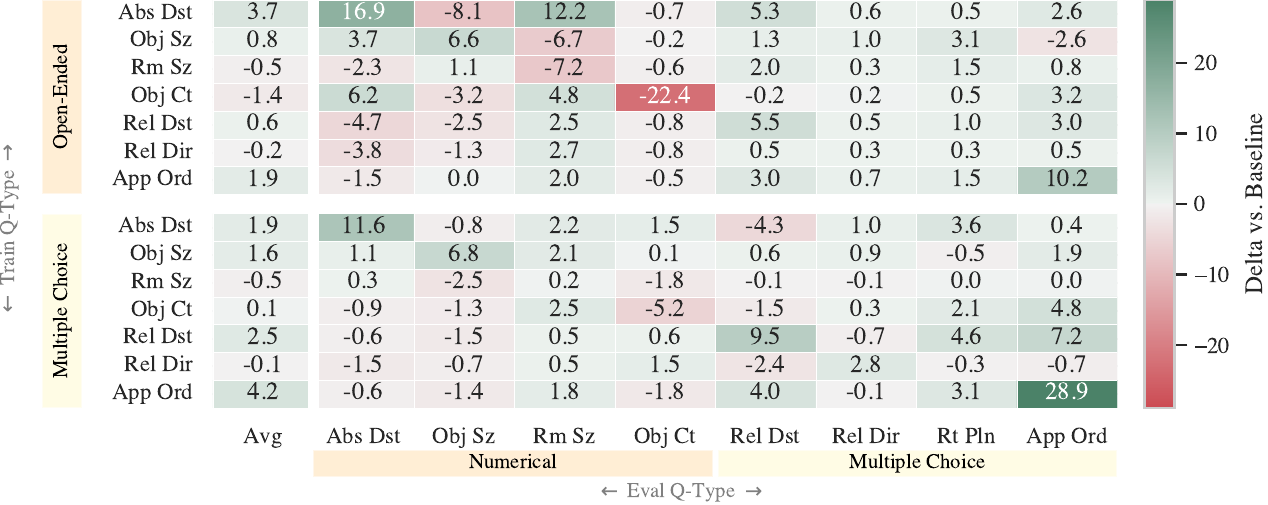}
    \caption{
        \textbf{Training on individual question types yields large on-task gains with localized cross-task effects.}
        We fine-tune LLaVA-Video-7B on 5k simulated questions of each question type and format (rows), evaluating each model on all VSI-Bench question types (columns).
        Values are performance $\Delta$ vs.\ the pretrained baseline (positive is green, negative is red).
    }\label{fig:question_type_iso}
\end{figure*}

To investigate what properties of simulated data are most important for developing real-world spatial reasoning capabilities, we conduct a systematic study examining question types, data mix design, and quantity requirements.
Using LLaVA-Video-7B~\cite{zhang2024video} with Qwen2~\cite{yang2024qwen2} as our base model, we fine-tune on various configurations of our \dataset{} data and evaluate on VSI-Bench~\cite{yang2024think}, a comprehensive benchmark that tests spatial intelligence using \emph{real-world} videos.

\subsection{Question Type Transfer Analysis}\label{sec:question_types}
We first investigate \emph{which individual types of spatial reasoning questions are most effective for developing transferable spatial intelligence in video language models}. Given the vast range of possible spatial questions that could be generated from simulator metadata, identifying which question formats provide the strongest learning signal is crucial for efficient training.

To establish a rigorous baseline for evaluating sim-to-real transfer, we generate questions that mirror the format and structure of those in the VSI-Bench evaluation set (see \cref{fig:question_type_visualization}). This ensures any performance gains can be directly attributed to the model's ability to leverage simulated data rather than differences in question formulation or evaluation criteria.
\footnote{We did not implement route planning questions due to its complexity.}
To systematically evaluate question effectiveness, we tune LLaVA-Video-7B on 5K examples of each question type separately, and then evaluate performance across all categories on the VSI-Bench test set. This approach allows us to measure both the direct impact of each question type on its corresponding benchmark category and any transfer effects to other spatial reasoning capabilities.

As shown in \cref{fig:question_type_iso}, training on a single type primarily boosts its \emph{own} category, with selective cross-task (off-diagonal) effects.
Among broader families, \emph{spatiotemporal} (appearance order) and \emph{measurement} (absolute distance) drive the largest improvements: appearance order yields MC:\ +28.9 and OE:\ +10.2 on its category, while absolute distance yields OE:\ +16.9 and MC:\ +11.6.
Cross-task gains are modest but interpretable, e.g., absolute distance $\to$ room size (OE:\ +12.2) and relative distance $\to$ appearance order / route planning (MC:\ +7.2 / +4.6). In contrast, object counting can hurt its own category (OE:\ $-22.4$, MC:\ $-5.2$), possibly due to greater object diversity in our simulated QAs compared to those in VSI-Bench.
These patterns suggest emphasizing temporal and metric cues over broad type coverage when constructing simulated supervision.

\subsection{Data Mix Design and Scaling}\label{sec:question_mixes}
Next, we explore how combining several question types can create more comprehensive training mixes that develop robust spatial understanding across multiple dimensions.

\paragraph{VSI-Baseline Mix}
As our first approach, we create a training mix that closely mirrors the VSI-Bench test set in both question types and distribution. This approach minimizes the gap between training and evaluation tasks, allowing us to evaluate the effect of sim-to-real transfer without other confounding factors such as differences in question phrasing.
However, now we wish to evaluate the mixes of question types that generalize to other question types as well in the VSI-Benchmark and other spatial benchmarks.

\paragraph{3Q Minimal Mix}
Building on insights from our individual question type analysis, we hypothesize that a minimal set of complementary question types could be sufficient to develop strong spatial reasoning capabilities, without needing to mirror the full VSI-Bench distribution.
We select three representative questions that cover the core dimensions of spatial reasoning:
\begin{itemize}[itemsep=0.125em,topsep=0pt,leftmargin=16pt]
    \item \emph{Measurement}: Absolute distance estimation (open-ended) tests the ability to perceive and estimate metric properties of space.
    \item \emph{Perspective}: Relative direction determination (multiple-choice) evaluates understanding of perspective-dependent spatial configurations.
    \item \emph{Spatiotemporal}: Appearance order tracking (multiple-choice) assesses the ability to track objects across time and remember their temporal relationships.
\end{itemize}
\noindent These question types complement each other, requiring the model to develop comprehensive spatial representations covering metric properties, directional relationships, and temporal dynamics.

\begin{figure}[t]
    \centering
    \begin{minipage}[t]{4.8in}
        \centering
        \vspace{0pt}
        \includegraphics[width=\linewidth]{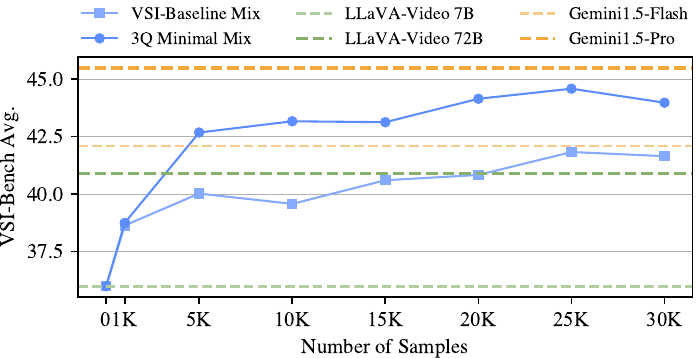}
    \end{minipage}%
    \hfill
    \begin{minipage}[t]{1.7in}
        \centering
        \vspace{9.05pt}
        \small
        \begin{tabular}{lr}
            \toprule
            \textbf{Q-Type} & \textbf{Pct} \\
            \hline
            \rowcolor{orange!10}
            \multicolumn{2}{l}{\textit{Open-Ended}} \\
            Abs Dist  & 16.9 \\
            Obj Size  & 19.3 \\
            Rm Size   & 5.8  \\
            Obj Ct    & 11.4 \\
            \hline
            \rowcolor{yellow!10}
            \multicolumn{2}{l}{\textit{Multi-Choice}} \\
            Rel Dist  & 14.4 \\
            Rel Dir   & 19.6 \\
            Rt Plan   & 0.0  \\
            App Ord   & 12.5 \\
            \bottomrule
        \end{tabular}

        \vspace{2em}
        \footnotesize Full Q-Type Dist.
    \end{minipage}
    \caption{
        \textbf{Minimal 3Q mix is more data-efficient than comprehensive coverage.}
        \textbf{\textit{Left:}} Both training mixes show rapid improvement on VSI-Bench, 
        with 3Q consistently outperforming the full baseline mix despite using 
        only three question types. At 5K examples, we surpass Gemini-1.5 Flash; 
        at 25K, we approach Gemini-1.5 Pro.
        \textbf{\textit{Right:}} Distribution of question types in VSI-Baseline mix, 
        which mirrors the VSI-Bench test set composition.
    }\label{fig:scaling_curve}
\end{figure}

\paragraph{How much data is needed?}
To understand the data efficiency of our approach and compare the two mixes, we scale the training set from 1K to 25K examples for both the VSI-Baseline and 3Q Minimal mixes.
As shown in \cref{fig:scaling_curve}, performance on VSI-Bench climbs rapidly for both mixes, with the 3Q Minimal mix reaching 42.5\% at just 5K examples---already surpassing Gemini-1.5 Flash (42.1\%).
Notably, the 3Q mix consistently outperforms the VSI-Baseline mix across all data scales, achieving 44.4\% at 25K examples and nearly matching Gemini-1.5 Pro (45.4\%), despite being trained on a more focused set of question types.
This demonstrates two key insights: \textbf{(1)} high-quality spatial annotations enable remarkably data-efficient learning, with strong performance achievable from just thousands of examples; and \textbf{(2)} a carefully selected minimal question set can be more effective than comprehensive coverage, suggesting that focused supervision on core spatial reasoning dimensions is sufficient to develop transferable capabilities.
Based on these findings, we use the 3Q Minimal mix for our main results and generalization experiments.

\begin{table}[p]
    \centering
    \caption{
        \textbf{VSI-Bench Performance: Competitive with large proprietary models.}
        LLaVA-Video-7B fine-tuned on 25K \framework{} examples achieves 44.4\%, surpassing 
        GPT-4o (34.0\%), the larger LLaVA-Video-72B (41.2\%), and approaching 
        Gemini-1.5 Pro (45.4\%). Strong gains in appearance order (+26.4\%) and 
        absolute distance (+20.0\%) demonstrate effective spatial transfer to real-world videos.
        LLaVA-OneVision-7B fine-tuned on the same set achieves similar gains.
    }\label{tab:vsib_orig}
    \resizebox{0.9\textwidth}{!}{
        \begin{tabular}{l|rrr|rrrrrrrr}
            &
                \rotatebox{75}{Macro} &
                \rotatebox{75}{Num} &
                \rotatebox{75}{MC} &
                \rotatebox{75}{Abs Dst} &
                \rotatebox{75}{Obj Sz} & 
                \rotatebox{75}{Rm Sz} &
                \rotatebox{75}{Obj Ct} &
                \rotatebox{75}{Rel Dst} &
                \rotatebox{75}{Rel Dir} &
                \rotatebox{75}{Rt Pln} &
                \rotatebox{75}{App Ord} \\
            Methods &
                \multicolumn{3}{c|}{\cellcolor{gray!10}Average} &
                \multicolumn{4}{c}{\cellcolor{orange!10}Numerical Answer} &
                \multicolumn{4}{c}{\cellcolor{yellow!10}Multiple-Choice Answer} \\
            \hline\\[-1.15em]
            \hline
            \rowcolor{navyblue!5}
            \multicolumn{1}{l|}{\textcolor{black}{\textit{Statistics}}} & & & & & & & & & & & \\
            Chance Level (Rand.) & {--} & {--} & 28.6 & {--} & {--} & {--} & {--} & 25.0 & 36.1 & 28.3 & 25.0 \\
            Chance Level (Freq.) & 34.0 & 39.3 & 31.7 & 32.0 & 29.9 & 33.1 & 62.1 & 25.1 & 47.9 & 28.4 & 25.2 \\
            \hline
            \rowcolor{navyblue!5}
            \multicolumn{1}{l|}{\textcolor{black}{\textit{Proprietary Models (API)}}} & & & & & & & & & & & \\
            GPT-4o & 34.0 & 33.4 & 34.6 & 5.3 & 43.8 & 38.2 & 46.2 & 37.0 & 41.3 & 31.5 & 28.5 \\
            Gemini-1.5 Flash & 42.1 & 47.1 & 37.0 & 30.8 & 53.5 & 54.4 & 49.8 & 37.7 & 41.0 & 31.5 & 37.8 \\
            Gemini-1.5 Pro & 45.4 & 48.7 & 42.1 & 30.9 & 64.1 & 43.6 & 56.2 & 51.3 & 46.3 & 36.0 & 34.6 \\
            Gemini-2.5 Pro & 51.5 & 46.5 & 56.5 & 34.9 & 64.3 & 42.8 & 43.8 & 61.1 & 47.8 & \bf 45.9 & 71.3 \\
            \hline
            \rowcolor{navyblue!5}
            \multicolumn{1}{l|}{\textcolor{black}{\textit{Open-source Models}}} & & & & & & & & & & & \\
            LLaVA-Video 72B       & 41.2 & \bf 42.4 & 40.0 & 24.5 & \bf 56.5 & 37.0 & \bf 51.4 & 41.7 & 36.1 & 33.0 & 49.2 \\
            LLaVA-Video 7B        & 36.0 & 34.0 & 38.0 & 15.2 & 46.9 & 24.1 & 49.7 & 44.1 & 42.4 & 33.5 & 31.9 \\
            \quad + 25k \framework{} 3Q   & \bf 44.4 & 40.3 & \bf 48.6 & \bf 35.2 & 41.2 & \bf 38.1 & 46.5 & \bf 53.8 & \bf 47.3 & \bf 35.1 & \bf 58.3 \\
            \rowcolor{inccol}
            \color{inctext}\quad $\Delta$ \textit{Improvement} & \p{+8.4} & \p{+6.3} & \p{+10.7} & \p{+20.0} & \n{-5.7} & \p{+14.0} & \n{-3.2} & \p{+9.7} & \p{+4.9} & \p{+1.6} & \p{+26.4} \\
            \hline
            LLaVA-OneVision 72B   & \bf 40.9 & \bf 42.5 & 39.3 & 25.2 & \bf 57.0 & \bf 41.8 & 46.1 & 42.8 & 34.9 & \bf 32.0 & 47.6 \\
            LLaVA-OneVision 7B    & 35.0 & 34.7 & 35.3 & 14.8 & 47.6 & 23.1 & \bf 53.2 & 43.4 & 38.4 & 31.4 & 27.8 \\
            \quad + 25k \framework{} 3Q   & 40.4 & 39.1 & \bf 41.8 & \bf 31.2 & 44.9 & 29.1 & 51.0 & \bf 44.4 & \bf 45.1 & 28.9 & \bf 48.9 \\
            \rowcolor{inccol}
            \color{inctext}\quad $\Delta$ \textit{Improvement} & \p{+5.4} & \p{+4.4} & \p{+6.6} & \p{+16.4} & \n{-2.7} & \p{+6.0} & \n{-2.2} & \p{+1.0} & \p{+6.7} & \n{-2.5} & \p{+21.1} \\
        \end{tabular}
    }
\end{table}

\begin{table}[p]
    \centering
    \caption{
        \textbf{VSI-Bench-Debiased Performance: Gains persist with reduced non-visual shortcuts.}
        Results on the debiased version~\cite{brown2025benchmark} show consistent 
        improvements (LLaVA-Video +7.7\%, LLaVA-OneVision +5.4\%), confirming that 
        \framework{} develops genuine visual reasoning rather than exploiting statistical 
        artifacts. Category-level patterns mirror the original benchmark.
    }\label{tab:vsib_deb}
    \resizebox{0.9\textwidth}{!}{
        \begin{tabular}{l|rrr|rrrrrrrr}
            &
                \rotatebox{75}{Macro} &
                \rotatebox{75}{Num} &
                \rotatebox{75}{MC} &
                \rotatebox{75}{Abs Dst} &
                \rotatebox{75}{Obj Sz} & 
                \rotatebox{75}{Rm Sz} &
                \rotatebox{75}{Obj Ct} &
                \rotatebox{75}{Rel Dst} &
                \rotatebox{75}{Rel Dir} &
                \rotatebox{75}{Rt Pln} &
                \rotatebox{75}{App Ord} \\
            Methods &
                \multicolumn{3}{c|}{\cellcolor{gray!10}Average} &
                \multicolumn{4}{c}{\cellcolor{orange!10}Numerical Answer} &
                \multicolumn{4}{c}{\cellcolor{yellow!10}Multiple-Choice Answer} \\
            \hline\\[-1.15em]
            \hline
            \rowcolor{navyblue!5}
            \multicolumn{1}{l|}{\textcolor{black}{\textit{Proprietary Models (API)}}} & & & & & & & & & & & \\
            Gemini-1.5 Flash & 39.9 & 44.8 & 35.0 & 28.5 & 45.7 & 56.2 & 48.9 & 35.5 & 34.7 & 28.2 & 41.4 \\
            Gemini-1.5 Pro   & 40.1 & 43.2 & 37.1 & 28.9 & 51.2 & 44.5 & 48.0 & 51.4 & 38.7 & 24.8 & 33.6 \\
            Gemini-2.5 Pro   & 49.1 & 43.2 & 55.1 & 34.4 & 56.7 & 40.9 & 40.8 & 61.0 & 46.8 & 43.0 & 69.5 \\
            \hline
            \rowcolor{navyblue!5}
            \multicolumn{1}{l|}{\textcolor{black}{\textit{Open-source Models}}} & & & & & & & & & & & \\
            LLaVA-Video 72B       & 36.8 & \bf 34.8 & 38.8 & 22.3 & \bf 45.8 & 33.7 & 37.3 & 43.9 & 37.8 & \bf 25.4 & 48.1 \\
            LLaVA-Video 7B        & 30.7 & 27.7 & 33.7 &15.4 & 34.7 & 22.1 & \bf 38.6 & 42.6 & 37.9 & 20.2 & 34.0 \\
            \quad + 25k \framework{} 3Q   & \bf 38.4 & 34.3 & \bf 42.6 & \bf 31.5 & 32.3 & \bf 37.6 & 35.7 & \bf 51.9 & \bf 39.2 & 19.3 & \bf 60.1 \\
            \rowcolor{gray!3}
            \color{inctext}\quad $\Delta$ \textit{Improvement} & \p{+7.7} & \p{+6.6} & \p{+9.0} & \p{+16.1} & \n{-2.4} & \p{+15.5} & \n{-2.9} & \p{+9.3} & \p{+1.3} & \n{-0.9} & \p{+26.1} \\
            \hline
            LLaVA-OneVision 72B   & \bf 35.6 & \bf 34.7 & 36.5 & 21.8 & \bf 44.3 & \bf 40.8 & 32.0 & 39.0 & 37.3 & 21.9 & 47.8 \\
            LLaVA-OneVision 7B    & 28.5 &25.7 & 31.3 & 16.3 & 31.4 & 18.0 & \bf 37.1 & 39.7 & 34.2 & \bf 23.7 & 27.7 \\
            \quad + 25k \framework{} 3Q   & 33.9 & 28.7 & \bf 39.1 &\bf 22.1 & 32.4 & 26.6 & 33.6 & \bf 43.9 & \bf 37.4 & 22.8 & \bf52.2 \\
            \rowcolor{gray!3}
            \color{inctext}\quad $\Delta$ \textit{Improvement} & \p{+5.4} & \p{+3.0} & \p{+7.8} & \p{+5.8} & \p{+1.0} & \p{+8.6} & \n{-3.5} & \p{+4.2} & \p{+3.2} & \n{-0.9} & \p{+24.5} \\
        \end{tabular}
    }
\end{table}

\section{Sim-to-Real Evaluation on Real-World Benchmarks}\label{sec:evaluation}

Having established that our 3Q Minimal mix provides the most effective and efficient training approach through controlled ablations, we now rigorously evaluate its ability to transfer spatial reasoning capabilities from simulation to real-world videos.
We train both LLaVA-Video-7B and LLaVA-OneVision-7B on 25K examples to assess sim-to-real transfer strength and architectural robustness.
While both models share the same backbone (SigLIP vision encoder and Qwen2-7B language model), they differ in their specialization: LLaVA-Video is optimized for video understanding with video-centric pretraining and temporal token allocation, while LLaVA-OneVision is a generalist model trained for single-image, multi-image, and video tasks.
This controlled comparison at identical parameter scale allows us to isolate whether improvements stem from our spatial training data itself rather than architecture-specific effects.

\subsection{Transfer to VSI-Bench Spatial Reasoning}\label{sec:vsibench_results}

Our training data was explicitly designed as a simulated ``train set'' for the VSI-Bench test set to investigate the potential of sim-to-real transfer for spatial reasoning in video.

\paragraph{Evaluation on VSI-Bench-Debiased.}
To ensure our results reflect genuine visual reasoning rather than exploitation of statistical shortcuts, we evaluate on both the original VSI-Bench and VSI-Bench-Debiased~\cite{brown2025benchmark}.
VSI-Bench-Debiased is a refined version that systematically reduces non-visual shortcuts through diagnostic analysis, providing a more rigorous test of visual understanding.
We present detailed results on both benchmarks: the original for comparison with proprietary models and prior work, and the debiased version to demonstrate robust visual reasoning capabilities.

Our 3Q Minimal mix model achieves 44.4\% on the original VSI-Bench (\cref{tab:vsib_orig}), surpassing GPT-4o (34.0\%) and approaching Gemini-1.5 Pro (45.4\%), despite being trained on just 25K examples.
The model shows particularly strong improvements in spatiotemporal (appearance order: +26.4\%) and measurement (absolute distance: +20.0\%) capabilities.
Critically, these improvements transfer consistently to VSI-Bench-Debiased (\cref{tab:vsib_deb}), where LLaVA-Video-7B achieves 38.4\% (+7.7\% over baseline) and LLaVA-OneVision-7B achieves 33.9\% (+5.4\% over baseline).
The consistent gains across both benchmark versions confirm that our approach develops genuine visual reasoning capabilities rather than exploiting statistical shortcuts, as the debiased version is specifically designed to reduce non-visual solvability.
Additionally, the strong performance of LLaVA-OneVision demonstrates that our findings are not architecture-specific but reflect fundamental principles of learning spatial reasoning from simulation.

\subsection{Generalization to Diverse Benchmarks}\label{sec:generalization}

To verify that spatial-focused training does not degrade general capabilities and that learned skills transfer beyond our target VSI-Bench evaluation, we assess performance across diverse tasks spanning video understanding, embodied reasoning, and real-world scenarios.

\paragraph{Maintained General Video Understanding.}
Our approach maintains stable performance on general video understanding benchmarks.
On VideoMME~\cite{fu2025videomme}, a comprehensive benchmark covering diverse real-world videos, performance remains essentially unchanged (LLaVA-Video: 63.3\%\textrightarrow{}63.1\%, LLaVA-OneVision: 58.3\%\textrightarrow{}59.0\%).
Similarly, EgoSchema~\cite{mangalam2023egoschema}, which tests long-form egocentric video understanding, shows modest improvements (+2.2\% for LLaVA-Video, stable for OneVision).
This demonstrates that spatial-focused tuning does not cause catastrophic forgetting of general video reasoning capabilities.

\paragraph{Transfer to Embodied and Real-World Spatial Tasks.}
Beyond maintaining general capabilities, our approach shows strong positive transfer to spatial tasks in new domains.
On OpenEQA~\cite{OpenEQA2023}, an embodied question-answering benchmark, LLaVA-Video improves dramatically (+8.6\%), demonstrating that spatial concepts learned in simulated indoor environments transfer to embodied reasoning scenarios.
Similarly, MMRealWorld~\cite{zhang2024mme}, which features high-resolution real-world outdoor images, shows consistent improvements (+4.5\% for LLaVA-Video, +1.1\% for OneVision), indicating that learned spatial understanding generalizes beyond the indoor training domain.

\paragraph{Architectural Robustness and Limitations.}
The consistent improvements across both LLaVA-Video (+8.4\% on VSI-Bench) and LLaVA-OneVision (+5.4\%) confirm that our findings reflect general principles of spatial learning from simulation rather than architecture-specific effects.
The larger gains for LLaVA-Video are expected given its video-centric pretraining and specialized temporal token allocation, while the solid improvements for the generalist LLaVA-OneVision model demonstrate that \framework{} training transfers effectively even to models not specifically optimized for video.
However, we observe modest negative transfer on some benchmarks, which we attribute to catastrophic forgetting from training exclusively on \dataset{} data without mixing with general instruction data.
Future work should explore optimal strategies for combining \dataset{} with broader multimodal training data to maximize spatial gains while preserving general capabilities.

\begin{table}[t]
    \centering
    \caption{
        \textbf{\framework{} training transfers to embodied and real-world scenarios 
        while maintaining general capabilities.}
        Results across diverse benchmarks show strong spatial transfer 
        (OpenEQA +8.6\%, MMRealWorld +4.5\% for LLaVA-Video) and stable egocentric (EgoSchema) and general video understanding (VideoMME) performance.
    }\label{tab:generalization}
    \resizebox{\textwidth}{!}{
        \begin{tabular}{l|cccccc}
            Model &
            \rotatebox{0}{VSI-B} &
            \rotatebox{0}{VSI-B\textsuperscript{Deb.}} &
            \rotatebox{0}{OpenEQA} &
            \rotatebox{0}{MME.RW\textsuperscript{lite}} &
            \rotatebox{0}{EgoSchema} &
            \rotatebox{0}{Video\textsuperscript{MME}} \\
    
        \hline\\[-1.15em]
        \hline

        \rowcolor{gray!10!blue!5} \textit{Proprietary Models} & & & & & & \\
        GPT-4o~\cite{hurst2024gpt}            & 34.0 & {--} & {--} & {--} & {--} & 71.9 \\
        Gemini-1.5-Pro~\cite{team2024gemini}    & 45.4 & 40.1 & {--} & {--} & 72.2 & 75.0 \\
        Gemini-2.5 Pro~\cite{comanici2025gemini}    & 51.5 & 49.1 & {--} & {--} & {--} & {--} \\
        
        \hline
        \rowcolor{gray!10!blue!5}\textit{Open-Source Models} & & & & & & \\
        LLaVA-Video 72B     & 41.2 & 36.8 & \bf 43.8 & 36.0 & \bf 66.7 & \bf 68.8 \\
        LLaVA-OneVision 72B & 40.9 & 35.6 & 43.0 & 48.2 & 62.8 & 66.7 \\
        \hline
        InternVL2.5 8B~\cite{chen2024expanding}      & 34.6 & 24.9 & {--} & {--} & 50.6 & 64.2 \\
        Qwen-VL-2.5 7B~\cite{Qwen2.5-VL}      & 33.5 & 29.6 & {--} & {--} & 65.0 & 65.1 \\
        \hline
        LLaVA-Video 7B      & 36.0 & 30.7 & 34.6 & 35.2 & 56.9 & 63.3 \\
        \quad + 25k \framework{} 3Q & \bf 44.4 & \bf 38.4 & 43.2 & 39.7 & 59.1 & 63.1 \\
        \rowcolor{gray!3}
        \color{inctext}\quad $\Delta$ \textit{Improvement} & \p{+8.4} & \p{+7.7} & \p{+8.6} & \p{+4.5} & \p{+2.2} & \n{-0.2} \\
        LLaVA-OneVision 7B  & 35.0 & 28.5 & 42.1 & 48.6 & 60.8 & 58.3 \\
        \quad + 25k \framework{} 3Q & 40.4 & 33.9 & 42.2 & \bf 49.7 & 60.5 & 59.0 \\
        \rowcolor{gray!3}
        \color{inctext}\quad $\Delta$ \textit{Improvement} & \p{+5.4} & \p{+5.4} & \p{+0.1} & \p{+1.1} & \n{-0.3} & \p{+0.7} \\
        \end{tabular}
    }
\end{table}

\section{Related Work}\label{sec:related_work}

Our work sits at the intersection of spatial reasoning in video-language models and synthetic data generation for multimodal training.

\paragraph{Spatial and 3D Reasoning in Vision-Language Models.}
Spatial understanding has been a long-standing challenge in computer vision, with early work focusing on grounding objects in 2D~\cite{peng2023kosmos, kamath2021mdetr, minderer2022simple} or 3D~\cite{chen2020scanrefer, hong20233d, achlioptas2020referit3d} using language descriptions.
Recent work addresses more complex spatial relationships between multiple objects~\cite{su202225d, liu2023visual, hudson2019gqa, ranasinghe2024learning}.
Despite advances in MLLMs~\cite{fu2024blink, chen2024spatialvlm, cheng2025spatialrgpt, tong2025cambrian, ramakrishnan2025does,yang2025cambrians}, spatial reasoning remains a fundamental weakness.
Recent approaches address this through pseudo-annotations on real images~\cite{cheng2025spatialrgpt} or synthetic data for single-image tasks~\cite{ray2025sat}, but video-based spatial reasoning with temporal dependencies~\cite{yang2024think} remains under-explored, as most video benchmarks emphasize action recognition~\cite{mangalam2023egoschema, miech2019howto100m} or embodied object recognition~\cite{OpenEQA2023} rather than spatiotemporal reasoning.

\paragraph{Video-Language Models and Datasets.}
Our work builds on multimodal foundation models~\cite{radford2021learning, jia2021scaling, wang2022omnivl, wang2022internvideo, singh2022flava} and recent video-language models~\cite{liu2023llava, zhu2024minigpt, openai2023gpt, team2023gemini, zhang2024llavanextvideo}.
While video-language datasets have evolved from static images~\cite{goyal2017making, hudson2019gqa} to temporal reasoning~\cite{fu2025videomme}, they predominantly focus on action recognition~\cite{mangalam2023egoschema, miech2019howto100m, das2013thousand, grauman2022ego4d} rather than geometric spatial relationships.
For example, questions like \textit{``what color is the cup?''} require only identifying the relevant frame, whereas we focus on spatiotemporal reasoning requiring metric distance estimation, perspective-taking, and tracking object appearances across time.

\paragraph{Learning from Synthetic Data for Spatial Understanding.}
Synthetic data has shown promise for improving spatial capabilities in vision-language models.
SAT~\cite{ray2025sat} uses simulation with action-based learning to improve spatial reasoning in static images, while SpatialRGPT~\cite{cheng2025spatialrgpt} employs pseudo-annotations on real images.
REVISION~\cite{chatterjee2024revision} introduces a 3D rendering pipeline to improve spatial fidelity in text-to-image \emph{generation} models, focusing on generating spatially accurate images from text prompts and evaluating MLLMs on spatial consistency.
Similarly, ``Going Beyond Nouns''~\cite{cascante2023going} uses the large-scale SyViC synthetic dataset to improve compositional understanding of attributes and relations in \emph{static images}.
Broader work in robotics demonstrates synthetic data's utility for manipulation~\cite{lin2025sim, matas2018sim} and navigation~\cite{ehsani2024spoc, zeng2025poliformer, silwal2024we}.
Our work differs in three key aspects:
\textbf{(1)} We focus on \emph{video-based spatiotemporal reasoning} for question-answering, requiring temporal memory and perspective-taking across minutes-long trajectories, rather than static image understanding or generation tasks;
\textbf{(2)} We systematically investigate which properties of simulated data (question types, data mixes, training scale) drive real-world transfer through controlled ablations; and
\textbf{(3)} We demonstrate strong empirical results, with our 7B model surpassing larger baselines and achieving competitive performance with proprietary models while showing robust generalization to embodied and real-world scenarios.
To enable these systematic investigations, we next describe our simulation-based data generation framework.

\section{Discussion}\label{sec:discussion}

\paragraph{Future Work and Limitations.}
We focus our investigation on spatial capabilities in video-language models using the LLaVA family.
Future work should explore whether our findings generalize to alternative recent architectures~\cite{grattafiori2024llama, deitke2024molmo, chen2025janus}.
Additionally, as noted in our generalization results, our experiments focus on fine-tuning exclusively on SIMS data without mixing with general instruction data, which can lead to some catastrophic forgetting.
Investigating optimal strategies for combining \framework{} with broader multimodal training data, and training at a larger data scale represent important avenues for maximizing spatial gains while preserving general capabilities.
An exciting direction is co-designing simulated training data with model-specific processing characteristics.
For instance, video-language models subsample frames during inference (e.g., 64 frames from 1800 in a 3-min video at 10 fps); ensuring that critical spatial information in the training samples remains visible across all possible subsampled frame sets could further improve learning efficiency.
The perfect ground truth of simulators enables such optimization---e.g., validating that each training example remains answerable under the target model's specific frame sampling strategy.

\paragraph{Conclusion.}
We propose \framework{}, a systematic framework for generating spatially-rich video training data from 3D simulators, and use it to investigate which properties of simulated data enable effective sim-to-real transfer.
Through controlled ablations, we identify minimal effective question types and demonstrate remarkable data efficiency, with just 5K--25K simulated examples enabling competitive performance with large proprietary models.
Our approach shows robust generalization to embodied and real-world scenarios while maintaining general video understanding capabilities.
We hope that our findings and released dataset pave the way for leveraging simulations as a scalable data source to improve spatial reasoning in video language models.

{
    \small
    \paragraph{Acknowledgments}
    We are grateful to the Ai2 Prior team for the excellent framework, codebase, and support that enabled this work.
    We thank Jihan Yang for helpful discussions and Shusheng Yang and Anjali Gupta for reviewing our manuscript.
    E.B. is supported by the DoD NDSEG Fellowship Program.
    S.X. acknowledges support from the MSIT IITP grant (RS-2024-00457882) and the NSF award IIS-2443404.
}

\clearpage
\bibliography{references}
\bibliographystyle{plain}

\clearpage
\appendix
\section*{Appendix}

This appendix provides comprehensive details about our experimental setup, dataset characteristics, and additional results, organized as follows:
\begin{itemize}[topsep=0pt]
    \item \S~\ref{app:implementation} details model architectures and training configurations
    \item \S~\ref{app:dataset_details} describes the \dataset{} dataset statistics, question templates, and prompt formats
    \item \S~\ref{app:additional_results} presents supplementary experimental analyses
\end{itemize}

\section{Implementation Details}\label{app:implementation}
We provide comprehensive details about our model architecture, training setup, and evaluation methodology to ensure reproducibility.

\subsection{Model Architecture}

We conduct experiments using two video-language models with identical backbones but different video specializations:

\paragraph{LLaVA-Video-7B~\cite{zhang2024llavanextvideo}}
\begin{itemize}
    \item Vision encoder: SigLIP-SO400M-patch14-384~\cite{zhai2023sigmoid}
    \item Language model: Qwen2-7B~\cite{yang2024qwen2}
    \item Video processing: Up to 64 frames per video with SlowFast-style temporal pooling
    \item Each frame encoded into a 12\texttimes{}12 grid of visual tokens
\end{itemize}

\paragraph{LLaVA-OneVision-7B~\cite{li2024llavaonevision}}
\begin{itemize}
    \item Vision encoder: SigLIP-SO400M-patch14-384~\cite{zhai2023sigmoid}
    \item Language model: Qwen2-7B~\cite{yang2024qwen2}
    \item Video processing: Up to 32 frames per video with per-frame token pooling
    \item Generalist model trained for single-image, multi-image, and video tasks
\end{itemize}

\noindent
For videos longer than the maximum frame count, we uniformly sample frames across the video duration.

\subsection{Training Configuration}
We fine-tune our models using the following hyperparameters:
\begin{itemize}
    \item Learning rate: 1e-6 (main and vision tower)
    \item Optimizer: AdamW (via DeepSpeed)
    \item Weight decay: 0.0
    \item Scheduler: Cosine with 3\% warmup
    \item Global batch size: 32
    \item Precision: Mixed precision (BF16)
    \item Gradient checkpointing: Enabled
    \item Maximum sequence length: 32,768 tokens
    \item Hardware: 8 $\times$ A100-80GB GPUs or 8 $\times$ H100 GPUs
\end{itemize}

For data scaling experiments, we maintain the same number of total training steps by adjusting the number of epochs for different dataset sizes (more epochs for smaller datasets).

\subsection{Data Processing}
For each simulated video, we:
\begin{itemize}
    \item Sample frames at 10 FPS
    \item Resize frames to 384 $\times$ 384 pixels
    \item Apply standard normalization based on the vision encoder's requirements
\end{itemize}

\section{\dataset{} Dataset Details}\label{app:dataset_details}

As a concrete instantiation of our framework, we release \dataset{}, comprising comprehensive spatial reasoning data across diverse simulated indoor environments.

\subsection{Dataset Statistics}
\dataset{} contains:
\begin{itemize}
    \item \textbf{203,048 total question-answer pairs}
    \item \textbf{2,507 unique video trajectories}
    \item \textbf{1,261 unique indoor scenes}
    \item \textbf{9 question types} in both open-ended and multiple-choice formats
\end{itemize}

\paragraph{Question Type Distribution.}
The dataset is balanced across question types and formats, with approximately 12.5k examples per question type per format (open-ended vs.\ multiple-choice), except for room size estimation which has approximately 2.5k examples per format due to its scene-level (rather than object-level) nature.
See \cref{tab:sims200k_breakdown} for the complete breakdown.

\begin{table}[h]
    \centering
    \small
    \caption{Per-question-type breakdown of \dataset{} dataset.}
    \label{tab:sims200k_breakdown}
    \begin{tabular}{lcc}
        \toprule
        Question Type & Open-Ended & Multiple-Choice \\
        \midrule
        Absolute Distance & 12,510 & 12,510 \\
        Appearance Order & 12,477 & 12,477 \\
        Object Count & 12,503 & 12,503 \\
        Relative Direction (Easy) & 12,250 & 12,250 \\
        Relative Direction (Medium) & 12,247 & 12,247 \\
        Relative Direction (Hard) & 12,230 & 12,230 \\
        Relative Distance & 12,295 & 12,295 \\
        Object Size Estimation & 12,503 & 12,503 \\
        Room Size Estimation & 2,509 & 2,509 \\
        \midrule
        \textbf{Total} & \textbf{101,524} & \textbf{101,524} \\
        \bottomrule
    \end{tabular}
\end{table}

\subsection{Question Templates}\label{app:question_templates}
Each question type follows a specific template that is populated with environment-specific details extracted from the simulator's metadata.
These templates create diverse, challenging tasks testing different aspects of spatial understanding, from metric estimation to perspective-taking and temporal tracking.

\begin{table*}[t]
    \centering
    \footnotesize
    \renewcommand{\arraystretch}{1.6}
    \setlength{\tabcolsep}{10pt}
    \caption{
        Question templates used in \framework{} to generate diverse spatial reasoning challenges. Each template creates questions that test a specific aspect of spatial understanding, from metric estimation to perspective-taking and temporal tracking.
    }\label{tab:question_templates}
    \begin{tabular}{lp{0.7\textwidth}}
        \toprule
        \textbf{Question Type} & \textbf{Template} \\
        \midrule
        
        Object Counting &
        How many \texttt{\{category\}}(s) are in this room? \\
        
        Object Size Estimation &
        What is the length of the longest dimension (length, width, or height) of the \texttt{\{category\}}, measured in centimeters? \\
        
        Room Size Estimation &
        What is the size of this room (in square meters)? If multiple rooms are shown, estimate the size of the combined space. \\
        
        Absolute Distance &
        Measuring from the closest point of each object, what is the direct distance between the \texttt{\{object1\}} and the \texttt{\{object2\}} (in meters)? \\
        
        Relative Distance &
        Measuring from the closest point of each object, which of these objects (\texttt{\{choice\_a\}}, \texttt{\{choice\_b\}}, \texttt{\{choice\_c\}}, \texttt{\{choice\_d\}}) is the closest to the \texttt{\{category\}}? If there are multiple instances of an object category, measure to the closest. \\
        
        Relative Direction (Hard) &
        If I am standing by the \texttt{\{positioning\_object\}} and facing the \texttt{\{orienting\_object\}}, is the \texttt{\{querying\_object\}} to my front-left, front-right, back-left, or back-right? The directions refer to the quadrants of a Cartesian plane (if I am standing at the origin and facing along the positive y-axis). \\
        
        Relative Direction (Med) &
        If I am standing by the \texttt{\{positioning\_object\}} and facing the \texttt{\{orienting\_object\}}, is the \texttt{\{querying\_object\}} to my left, right, or back? An object is to my back if I would have to turn at least 135 degrees in order to face it. \\
        
        Relative Direction (Easy) &
        If I am standing by the \texttt{\{positioning\_object\}} and facing the \texttt{\{orienting\_object\}}, is the \texttt{\{querying\_object\}} to the left or the right of the \texttt{\{orienting\_object\}}? \\
        
        Spatiotemporal Distance &
        Which of these objects (\texttt{\{choice\_a\}}, \texttt{\{choice\_b\}}, \texttt{\{choice\_c\}}, \texttt{\{choice\_d\}}) is the closest to the ego-position at the last frame in the video? \\
        
        Appearance Order &
        What will be the first-time appearance order of the following categories in the video: \texttt{\{choice\_a\}}, \texttt{\{choice\_b\}}, \texttt{\{choice\_c\}}, \texttt{\{choice\_d\}}? \\
        
        Route Planning &
        You are a robot beginning at the \texttt{\{start\_obj\}} and facing the \texttt{\{orienting\_obj\}}. You want to navigate to the \texttt{\{end\_obj\}}. You will perform the following actions (Note: for each [please fill in], choose either `turn back,' `turn left,' or `turn right.'): \texttt{\{actions\}} You have reached the final destination. \\
        
        \bottomrule
    \end{tabular}
\end{table*}

\subsection{Prompt Templates}
We use standardized prompts to format questions during training:

\begin{table}[h]
    \centering
    \small
    \caption{Prompting templates combined with question templates.}
    \label{tab:prompt_templates}
    \begin{tabular}{lp{0.6\textwidth}}
        \toprule
        \textbf{Prompt Type} & \textbf{Template} \\
        \midrule
        Video Introduction (pre-prompt) & These are frames of a video. \\
        Multiple-Choice (post-prompt) & Answer with the option's letter from the given choices directly. \\
        Numerical Answer (post-prompt) & Please answer the question using a single word or phrase. \\
        \bottomrule
    \end{tabular}
\end{table}

\section{Additional Results and Analysis}\label{app:additional_results}

This section provides supplementary experimental results that complement the main paper.

\subsection{Question Type Transfer on VSI-Bench-Debiased}

To verify that the question type transfer patterns observed in \cref{fig:question_type_iso} (main paper) are not artifacts of statistical shortcuts, we repeat the analysis on VSI-Bench-Debiased.

\begin{figure*}[t]
    \centering
    \includegraphics[width=\linewidth]{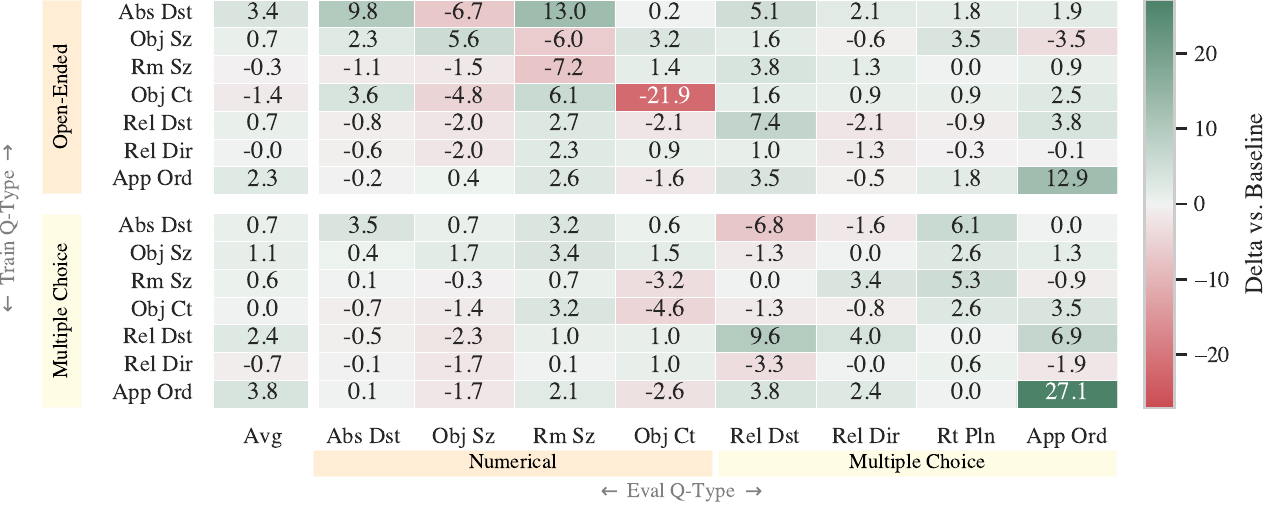}
    \caption{
        \textbf{Question type transfer patterns remain consistent on VSI-Bench-Debiased.}
        Similar to \cref{fig:question_type_iso}, appearance order and absolute distance show strong cross-task transfer on the debiased benchmark, confirming that observed patterns reflect genuine spatial learning rather than exploitation of statistical shortcuts.
    }\label{fig:question_type_iso_debiased}
\end{figure*}

\end{document}